\documentclass{article}

\usepackage{amsmath}
\usepackage{amssymb}
\usepackage{subcaption} 

\usepackage[dblblindworkshop, final]{neurips_ACA_2025}

\workshoptitle{\\Workshop on Algorithmic Collective Action (ACA@NeurIPS 2025)}

\usepackage[utf8]{inputenc} 
\usepackage[T1]{fontenc}    
\usepackage{hyperref}       
\usepackage{url}            
\usepackage{booktabs}       
\usepackage{amsfonts}       
\usepackage{nicefrac}       
\usepackage{microtype}      
\usepackage{listings}
\usepackage{xcolor}         
\usepackage{graphicx}

\lstdefinestyle{promptstyle}{
  basicstyle=\ttfamily\footnotesize,
  frame=single,
  backgroundcolor=\color{gray!9},
  breaklines=true,
  columns=fullflexible,
  keepspaces=true,
  showstringspaces=false
}

\title{Conscious Data Contribution via Community-Driven Chain-of-Thought Distillation}

\author{%
    Lena Libon $^{1}$\thanks{Correspondence to: \texttt{llibon@ethz.ch}} \quad
    Meghana Bhange $^{2}$ \quad
    Rushabh Solanki $^{3}$ \quad
    Elliot Creager $^{3}$  \quad
    Ulrich A\"ivodji $^{2}$ \\[1ex]    
    $^{1}$ETH Zurich \\
    $^{2}$\'ETS Montr\'eal, Mila \\
    $^{3}$ University of Waterloo, Vector Institute
}

\begin{document}
\maketitle

\begin{abstract}
   The current era of AI development places a heavy emphasis on training large models on increasingly scaled-up datasets. This paradigm has catalyzed entirely new product categories, such as LLM chatbots, while also raising concerns about data privacy and consumer choice. 
  In this paper, we consider questions of data portability and user autonomy in the context of LLMs that ``reason'' using chain-of-thought (CoT) traces, computing intermediate text artifacts from user input before producing a final output. 
  We first interpret recent data privacy and portability law to argue that these intermediate computations qualify as users' personal data. 
  Then, building on the existing framework of Conscious Data Contribution, we show how communities who receive low utility from an available model can aggregate and distill their shared knowledge into an alternate model better aligned with their goals.
  We verify this approach empirically and investigate the effects of community diversity, reasoning granularity, and community size on distillation performance. 
\end{abstract}
\section{Introduction}
As AI-based decision-making becomes increasingly embedded in our society, the question of how to make responsible AI development more participatory is gaining prominence. Traditional ``repair-from-above'' strategies, such as platform-led reforms or regulatory interventions, face several limitations, including slow-moving regulatory processes~\citep{gowlingwlg_billc27_timeline_2024}, industry engineering practices that often misalign with community needs~\citep{delgado2023participatory}, and the persistent risk of ethics washing~\citep{schultz2024digital}. Data leverage~\citep{vincent2021dataleverageframeworkempowering} has emerged as a promising approach to empower affected communities by enabling them to use their unique position as data generators to regain influence over data-dependent AI platforms. An important instantiation of this idea is Conscious Data Contribution (CDC)~\citep{10.1145/3449177}, which offers a means for affected parties to protest by contributing their data to competing organizations that align with their values. Real world examples of CDC often include migration between social media platforms, e.g. Twitter/X to Bluesky~\citep{cooper_bluesky_2025}, Reddit to Voat and Snapzu~\citep{newell_user_2016}, or WhatsApp users moving to Telegram and Signal due to change in privacy terms~\citep{noauthor_why_2021}. These can happen organically or organized by a pre-existing advocacy group.

Existing research on CDC in the context of machine learning has primarily conceptualized it in terms of contributions to training datasets and studied its impact through learning curve simulations~\citep{10.1145/3449177}. In these studies, a competing model is trained on various subsets of the original training data, representing different participation rates, and its resulting accuracy is used as a proxy for CDC’s effectiveness. These approaches assume a non-rivalry principle~\citep{jones2020nonrivalry}. Although valuable, this perspective overlooks an increasingly relevant form of user-generated data: interaction data produced during real-time engagements with AI systems.

This is particularly important because recent data protection frameworks recognize interaction data as a form of personal data. Under the European Union’s General Data Protection Regulation (GDPR) \citep{EU2016GDPR} and Quebec’s Act Respecting the Protection of Personal Information in the Private Sector (Loi 25) \citep{QuebecP391}, individuals are granted the right to inspect, control and transfer personal data. 
GDPR Article 4(1) defines \emph{personal data} as any information relating to an identified or identifiable natural person, including explicit identifiers (e.g., names and addresses) and indirect identifiers (e.g., behavioral profiles and communication patterns). Similarly, Section 2 of Quebec's Privacy Act defines personal information as any data that allow an individual to be identified, directly or indirectly. Both frameworks extend protection to data derived from user activity. For instance, GDPR Article 4(4) defines \emph{profiling} as any automated processing of personal data used to evaluate or predict certain personal aspects related to a natural person, with an analogous provision in Section 8.1. in Quebec's law. Consequently, in the context of conversational AI, both user-provided inputs (e.g., prompts) and AI-generated responses, including intermediate chain-of-thought (CoT) reasoning traces, may qualify as protected personal data. Importantly, these legal frameworks explicitly recognize the right to \emph{data portability}, as outlined in GDPR Article 20 and Section 27 of Quebec's law, allowing individuals to receive their personal data in a structured, machine-readable format and to transmit it to another provider.

Building on these legal foundations, we implement the CDC paradigm. When users contribute their personal data from the conversational history, including prompts, responses, and CoT reasoning, they could enable alternative AI providers to distill the knowledge from the original model. However, the effectiveness of this strategy, particularly when contributions come from multiple heterogeneous communities with varying interests, remains largely unexplored.

Our work addresses this gap through three key contributions. First, we present the \textbf{first systematic study of multi-community knowledge distillation under the CDC paradigm}, examining how interaction data can be leveraged to train alternative AI systems. Second, we uncover \textbf{nontrivial interactions between community diversity and reasoning granularity}, showing that more data or more detailed reasoning does not always translate to improved performance. Finally, we analyze \textbf{strategies for effectively combining heterogeneous community contributions}, offering practical insights into how participatory mechanisms can shape the future of responsible AI development. 

\section{Background and Related Work}
\paragraph{Knowledge Distillation}
Knowledge distillation is a widely used technique for transferring knowledge from a large teacher model to a smaller student model while maintaining comparable performance \citep{hinton2015distillingknowledgeneuralnetwork}. Traditional distillation typically assumes centralized, homogeneous datasets and relies on soft-label supervision using the teacher's output logits. However, recent advances have demonstrated that incorporating richer intermediate signals improves the reasoning ability of the student model. In particular, CoT-based distillation has emerged as an effective strategy. The Distilling Step-by-Step framework \citep{hsieh2023distillingstepbystepoutperforminglarger} shows that including intermediate reasoning steps, rather than only final answers, leads to better performance. However, existing CoT-based distillation approaches often assume centralized and homogeneous datasets, where all samples come from the same distribution. This assumption overlooks practical scenarios where data originates from multiple heterogeneous sources, such as communities with distinct question styles and domains.

An exception is MoDE-CoTD \citep{li-etal-2024-mode}, which goes beyond strict homogeneity by using a mixture of decoupled LoRA experts to distill reasoning abilities across diverse reasoning tasks. Freezing the base model and modularizing updates mitigates catastrophic forgetting and enables effective adaptation to unseen tasks. However, MoDE-CoTD addresses technical task diversity rather than the participatory dynamics of multi-community settings, in which data contributions are voluntary, heterogeneous and influenced by different objectives and constraints.

In contrast, our work examines knowledge distillation in a multi-community context inspired by CDC, where contributions are voluntary and strategically motivated. Unlike prior work, we explicitly analyze how heterogeneous communities with different interests interact under varying collective strategies: greedy, utilitarian, and altruistic (see \autoref{subsec:setting}).

\paragraph{Reasoning at Different Levels of Granularity}
The existing CoT literature suggests the complex and often counter-intuitive relationship between reasoning granularity and model performance.
The intuition that more detailed reasoning is always beneficial has been increasingly challenged.
For instance, \citet{chen2025unveilingkeyfactorsdistilling} demonstrates that Small Language Models (SLMs) exhibits a non-monotonic relationship with reasoning granularity.
Their study shows that fine-grained reasoning steps enhance the performance of stronger student models, whereas weaker models are often overwhelmed and instead perform better with simpler CoT supervision.

\citet{wu2025lessunderstandingchainofthoughtlength} further show that task accuracy often follows an inverted U-shaped curve with respect to CoT length, where performance degrades after an optimal number of steps.
Complementing this, other research has shown that lengthening reasoning steps, even without adding new semantic information, can significantly improve LLM performance, implying the importance of the reasoning process itself \citep{jin2024impactreasoningsteplength}.
Given prior findings that reasoning granularity influences model performance in different ways, we include multiple levels of granularity in our experiments.
Specifically, we evaluate both concise, high-level rationales and more detailed step-by-step breakdowns.

\section{Methodology}
\label{subsec:setting}
We study a multi-community knowledge distillation setting in which four distinct communities, represented by AQuA-RAT (AQuA)~\citep{ling-etal-2017-program}, CommonsenseQA (CSQA)~\citep{talmor-etal-2019-commonsenseqa}, OopenBookQA (OBQA)~\citep{mihaylov2018suitarmorconductelectricity}, and StrategyQA (STQA)~\citep{geva2021didaristotleuselaptop}, contribute their data to distill a larger teacher model into a smaller student model. These communities differ in domain and question formulation, reflecting the heterogeneity typical of real-world participatory data ecosystems. To make the setting tractable while preserving diversity, our experiments focus on pairwise combinations of these communities, enabling controlled analysis of cross-community transfer under variations in community composition.

We use LLaMA 3 70B as the teacher model for generating supervision signals and T5-base as the student model. For CoT-distillation, we adopt the Distilling Step-by-Step framework \citep{hsieh2023distillingstepbystepoutperforminglarger}. To examine the effect of different supervision signals, we compare two modes: answer-only, where the student is trained solely on final answers, and CoT-based, where reasoning chains accompany the answers during distillation. Within the CoT-based setting, we further investigate the role of reasoning granularity by following the taxonomy proposed by \citet{chen2025unveilingkeyfactorsdistilling}. Specifically, we compare concise and highly summarized reasoning traces (Level 1) with long and fine-grained explanations (Level 6), as illustrated in \autoref{appendix:level_granularity}. We also include an additional condition with summarized chains, reflecting realistic constraints where contributors might limit reasoning detail for privacy or cost considerations. The summarization process is described in \autoref{appendix:summarize_chains}.

To study scalability and participation effects, we conduct learning curve simulations by varying the size of contributing collectives, allowing us to observe how performance evolves as more data is shared. Finally, to better understand the dynamics of cross-community learning, we measure community diversity in terms of heterogeneity in reasoning styles and question formulations, and relate this diversity to agreement with the teacher model. This analysis enables us to assess whether higher diversity undermines consensus, or whether structured reasoning traces mitigate such effects. Details of the diversity metric are provided in \autoref{sec:diversity}.

Beyond supervision strategies, we consider how the structure of collective incentives influences contribution patterns. We model three strategic objectives representing different principles of cooperation: maximizing the accuracy of one’s own community (greedy), maximizing the weighted average accuracy across all communities (utilitarian) and maximizing the minimum accuracy to protect the least advantaged group (altruistic). These objectives capture fundamental trade-offs between self-interest and collective welfare that arise in participatory machine learning.

\section{Results and Analysis}
We analyze the impact of different design choices on the collective distillation performance. We focus primarily on utilitarian and altruistic objectives. These objectives represent fairness-efficiency trade-offs that are central to collaborative AI training. 
Finally, we introduce an additional perspective, the greedy strategy, which evaluates incentives for individual communities to strategically decide their level of contribution.
We also explore the impact of relative community size on distillation outcomes in~\autoref{app:rq4}. 

\subsection{RQ1: Impact of CoT on Distillation Performance}
\begin{figure}[h!]
    \centering
    \begin{subfigure}[t]{0.54\linewidth}
        \centering
        \includegraphics[width=\linewidth]{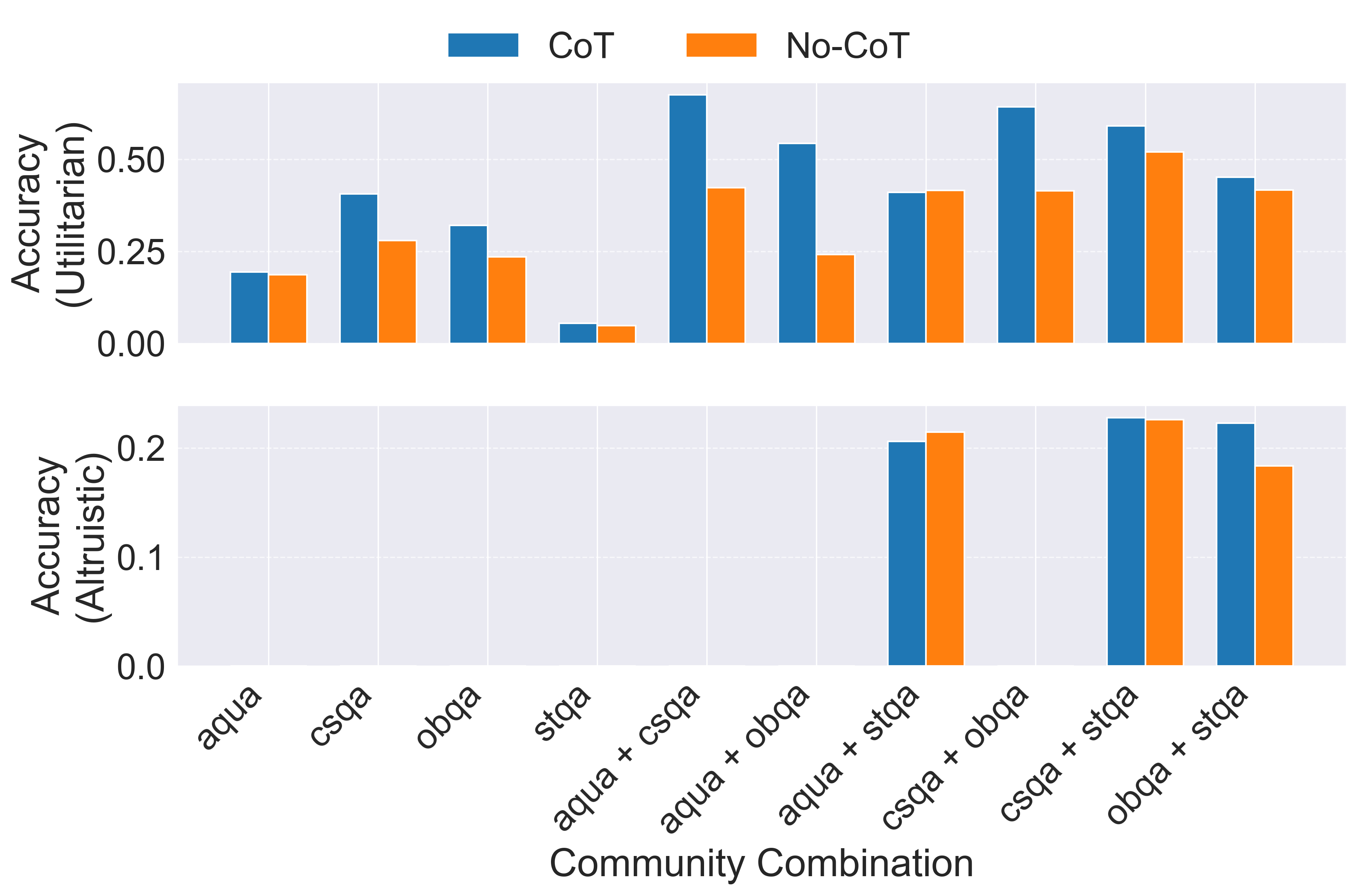}
        \caption{Impact of CoT on distillation performance across dataset combinations measured by the utilitarian and altruistic accuracy.}
    \label{fig:rq1}
    \end{subfigure}
    \hfill
    \begin{subfigure}[t]{0.43\linewidth}
        \centering
        \includegraphics[width=\linewidth]{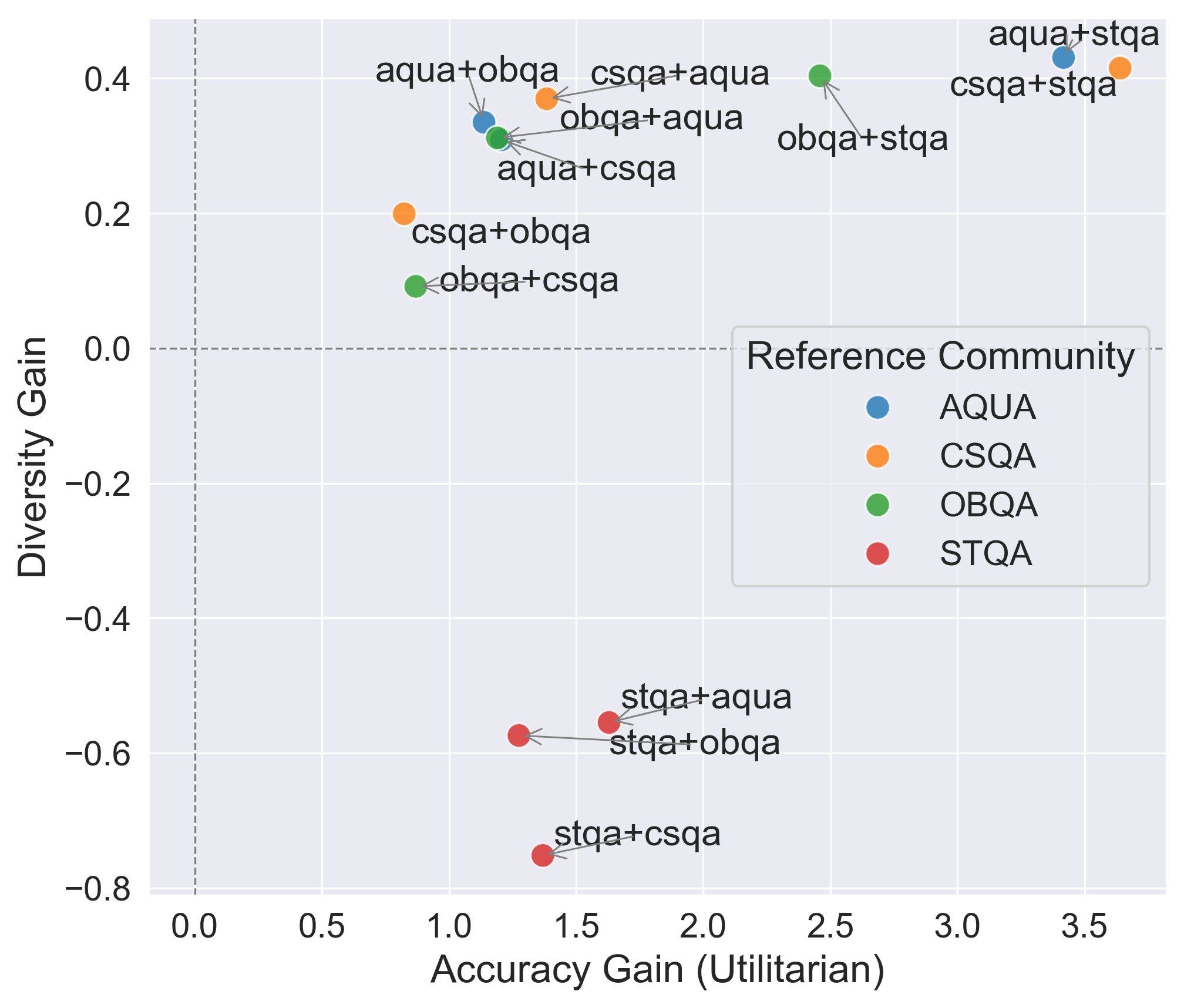}
        \caption{Diversity gain vs. accuracy gain for all two-community combinations under the utilitarian objective. Each point represents a directional combination (reference dataset $\rightarrow$ added dataset), illustrating asymmetric contributions to accuracy and diversity.}
        \label{fig:rq2}
    \end{subfigure}
    \caption{Results for RQ1 and RQ2.}
\end{figure}

We compare models distilled with and without CoT reasoning traces across all dataset combinations. \autoref{fig:rq1} presents performance for both utilitarian and altruistic objectives.

Under the utilitarian objective, the inclusion of CoT substantially improves performance on CSQA and OBQA, both when trained in isolation and in combination with other datasets. This indicates that tasks that require complex common sense reasoning and understanding of the topic benefit significantly from explicit reasoning traces. In contrast, for AQuA and STQA, the differences between CoT and no-CoT are minimal. When examining the altruistic objective, we observed many instances with zero accuracy, which can largely be attributed to differences in task formats between communities. STQA is a True/False dataset, while AQuA, CSQA, and OBQA are multiple-choice. Models distilled exclusively on multiple-choice datasets fail on True/False tasks, and vise versa. In combinations that include both formats, altruistic performance stabilizes, and the differences between CoT and no-CoT become minimal, except for the OBQA + STQA pairing, where CoT provides a slight advantage.

These results suggest that for a utilitarian collective, CoT is particularly beneficial when reasoning-intensive tasks, such as CSQA, are included in the coalition, as it enhances the average accuracy across all communities. In contrast, for an altruistic collective, the primary limiting factor is task-type coverage rather than the presence of reasoning traces. Ensuring diversity in task formats is, therefore, more critical than adding CoT for improving the minimum performance.

\subsection{RQ2: Effect of Community Diversity}  
We investigate whether combining multiple communities improves generalization or introduces conflicting signals. To quantify this, we compute the accuracy gain, defined as the improvement in utilitarian performance relative to the standalone performance of the added community, and the diversity gain, measured as the relative increase in diversity (measured by VendiScore~\cite{friedman_vendi_2023}) compared to the standalone diversity score of the added community, as described in \autoref{sec:diversity}. In this analysis, we focus exclusively on the utilitarian objective, since, as observed in RQ1, many accuracy values under the altruistic objective are zero due to format mismatches, making gains unreliable and potentially misleading. 

\autoref{fig:rq2} visualizes the relationship between the accuracy gain and the diversity gain in all dataset combinations. The scatter plot reveals a clear positive correlation for all cases where STQA is not the reference dataset. In these cases, the diversity gain is consistently positive, indicating that adding another community improves both accuracy and diversity simultaneously. This suggests that combining multiple-choice datasets, such as AQuA, CSQA, and OBQA, generally leads to mutually beneficial contributions, enhancing both generalization and the diversity of knowledge represented in the model. In contrast, when STQA is the reference dataset, the pattern changes. Here, diversity gains are negative, and no positive correlation is observed. Accuracy gains range approximately from $1.35$ to $1.65$, but added communities do not contribute positively to diversity. This effect likely stems from the task format mismatch: When STQA serves as the reference, adding a multiple-choice dataset can improve utilitarian accuracy without increasing diversity because the added dataset contributes little to the representation of STQA-specific task characteristics. 

These observations underscore the importance of considering both dataset compatibility and task format when forming multi-community coalitions. Positive, mutually reinforcing gains are most likely when communities share similar task structures, while heterogeneous combinations can produce asymmetries in diversity contribution despite improvements in average accuracy.

\subsection{RQ3: Impact of Reasoning Granularity} 
We further explore how the level of detail in the reasoning traces affects the distillation performance. We compare four configurations: detailed CoTs (level 6), minimal CoTs (level 1), and summarized reasoning traces generated by models of two different capacities (8B and 70B). For consistency, this analysis focuses exclusively on the utilitarian objective, as altruistic performance remains heavily influenced by format incompatibilities, which obscure the impact of reasoning granularity.

\begin{figure}[h!]
    \centering
    \begin{subfigure}[t]{0.55\linewidth}
        \centering
        \includegraphics[width=\linewidth]{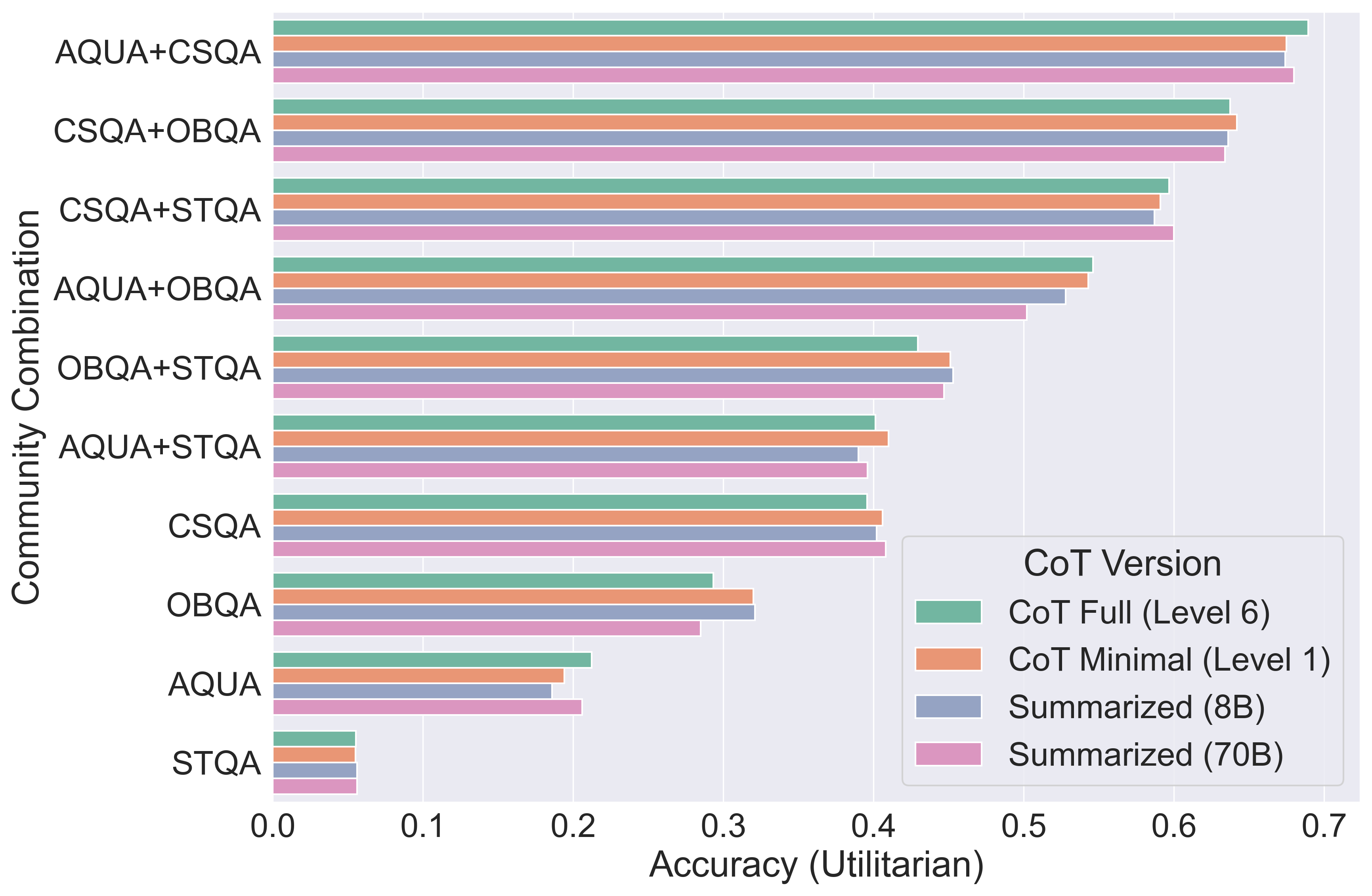}
        \caption{Comparison of utilitarian accuracy across dataset combinations for different reasoning granularity levels: full CoT (level 6), minimal CoT (level 1), and summarized CoT traces generated by 8B and 70B models.}
        \label{fig:rq3}
    \end{subfigure}
    \hfill
    \begin{subfigure}[t]{0.43\linewidth}
        \centering
        \includegraphics[width=\linewidth]{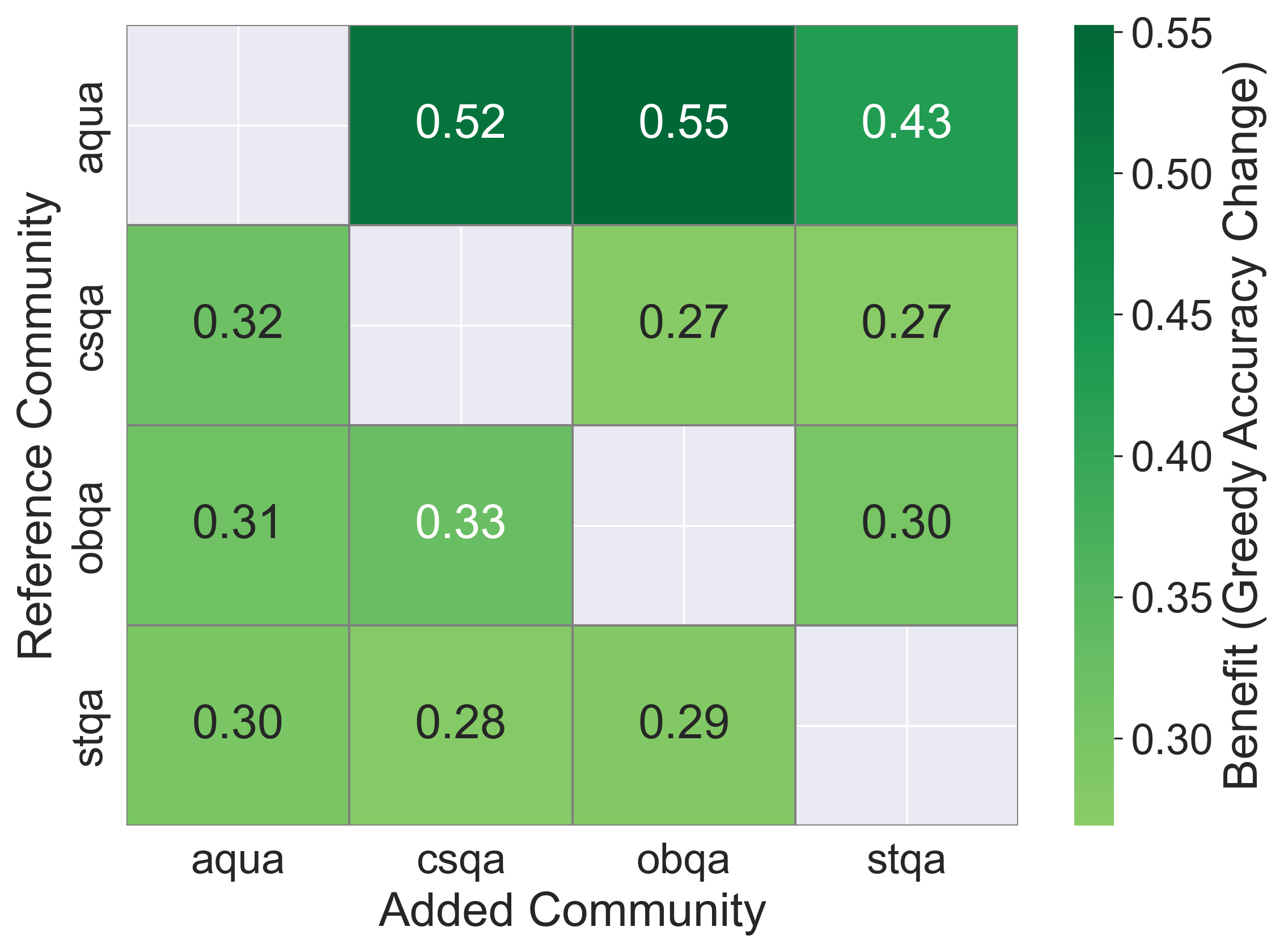}
        \caption{Greedy benefit: accuracy gain for each community when paired with another. Positive values indicate benefit from joint distillation.}
        \label{fig:rq5}
    \end{subfigure}
    \caption{Results for RQ3 and the greedy perspective.}
\end{figure}

\autoref{fig:rq3} compares accuracy across all dataset combinations under these four reasoning configurations. Contrary to expectations, the plot does not reveal a strong or consistent advantage for detailed reasoning traces over more concise or summarized alternatives. The differences across configurations are generally small and lack a clear systematic trend. One plausible explanation is that once a minimal level of reasoning is provided, additional detail neither enhances nor degrades performance in a meaningful way.

\subsection{Greedy Perspective: Strategic Contribution Analysis}
In this analysis, we adopt the perspective of individual communities acting strategically, focusing solely on maximizing their own performance rather than optimizing global fairness or average utility. We define benefit as the change in a community’s accuracy on its own dataset when participating in a joint distillation process, compared to training independently on that community alone. A positive benefit indicates that the community benefits from collaboration, whereas a negative value would imply a loss relative to solitary training. The heatmap in \autoref{fig:rq5} summarizes these benefits for all pairs of communities. Each row corresponds to a reference community, and each column represents an added community in the coalition. Diagonal cells are left empty because a community cannot add itself. The values in the off-diagonal cells capture the increase in greedy accuracy resulting from the joint distillation. The heatmap shows that all reference communities experience positive benefits when collaborating with other communities, indicating that joint distillation is generally advantageous. AQuA, which consists of mathematics questions, exhibits the highest gains, even when combined with non-math datasets such as OBQA, achieving an accuracy improvement of $0.55$. This suggests that communities can extract complementary knowledge from datasets with different task characteristics. Similarly, STQA benefits from all added communities, despite its True/False format differing from the multiple-choice nature of the others. These results highlight that the utility of collaboration is not strictly constrained by task format and that even seemingly heterogeneous datasets can provide valuable information for individual communities.

Overall, this analysis underscores that communities acting in their own interest can still benefit from cooperation. The magnitude of benefit depends on the reference community and the added dataset, reflecting asymmetries in knowledge contribution. 

\section{Discussion}
\subsection{Practical Implications for CDC}
CDC success depends on balancing collective utility with individual incentives. Reasoning traces offer the greatest value for reasoning-intensive datasets such as CSQA and OBQA, while their impact on more structured tasks such as AQuA is limited. Importantly, even summarized CoT chains deliver substantial gains for complex reasoning tasks, indicating that full reasoning traces are not strictly required to improve performance. However, benefits are uneven: For example, AQuA gains significantly more than others. This asymmetry may undermine the willingness to collaborate unless mechanisms for fair benefit sharing or strategic coalition design are introduced. 

\subsection{Limitations and Future Work}

\paragraph{Robustness to Platform Countermeasures} While our current work focuses on communities participating in CDC, platforms may also notice the trend of users moving away and may employ countermeasures. This could involve using techniques such as antidistillation sampling ~\citep{savani_antidistillation_2025}. Understanding how CDC performs under conditions in which the platform actively employs countermeasures could help communities gain insight into the robustness of these techniques. 

\paragraph{Expansion to Real-World Datasets} Benchmark datasets such as AQuA-RAT, CSQA, OBQA, and STQA provide controlled evaluation settings but do not capture the diversity and complexity of real participatory data. In practice, questions vary widely in style, scope, and formulation, both within and between communities. Real-world datasets naturally include open-ended, multistep, or ambiguously phrased questions, which can also vary in task format (e.g., multiple choice, short answer). Future work could construct such datasets by collecting community Q\&A threads or collaborative problem-solving platforms, preserving the natural variety of questions and contributor behavior to create more realistic and heterogeneous conditions.

\paragraph{Costs and Incentives of Cooperation} Our work shows how various interactions between different communities can lead to varying benefits when coming from an altruistic or utilitarian perspective. This parallels to different settings studies in economic and game-theoretic domains, for instance, Public Goods Game. Expanding the work through the lens of cooperative games for communities might help to understand conditions under which incentives for communities to cooperate, trade-offs, and possible free-riding effect when contributing data.

\section*{Acknowledgements}
The resources used in preparing this research were provided, in part, by the Province of Ontario, the Government of Canada through CIFAR, and companies sponsoring the Vector Institute~\url{www.vectorinstitute.ai/partnerships/}. The authors thank the Digital Research Alliance of Canada for computing resources. Ulrich Aïvodji is supported by NSERC Discovery grant (RGPIN-2022-04006) and IVADO's Canada First Research Excellence Fund to develop Robust, Reasoning and Responsible Artificial Intelligence (R$^3$AI) grant (RG-2024-290714).

\bibliographystyle{plainnat}
\bibliography{citations.bib}


\appendix
\section{Additional Experiments: Community Size Dynamics} \label{app:rq4}
We examine how the relative size of communities influences distillation outcomes. For this analysis, we focus on the combination of CSQA and STQA. This pair was chosen for two reasons: first, it exhibited the highest accuracy gain and one of the highest diversity gains in \autoref{fig:rq2}, making it an exemplary case of synergistic collaboration. Second, CSQA and STQA differ substantially in format. CSQA is a multiple choice dataset, while STQA follows a True/False structure, which offers an interesting testbed for understanding how size imbalances affect performance in heterogeneous settings.

\begin{figure}[h!]
    \centering
    \begin{subfigure}[t]{0.46\linewidth}
        \centering
        \includegraphics[width=\linewidth]{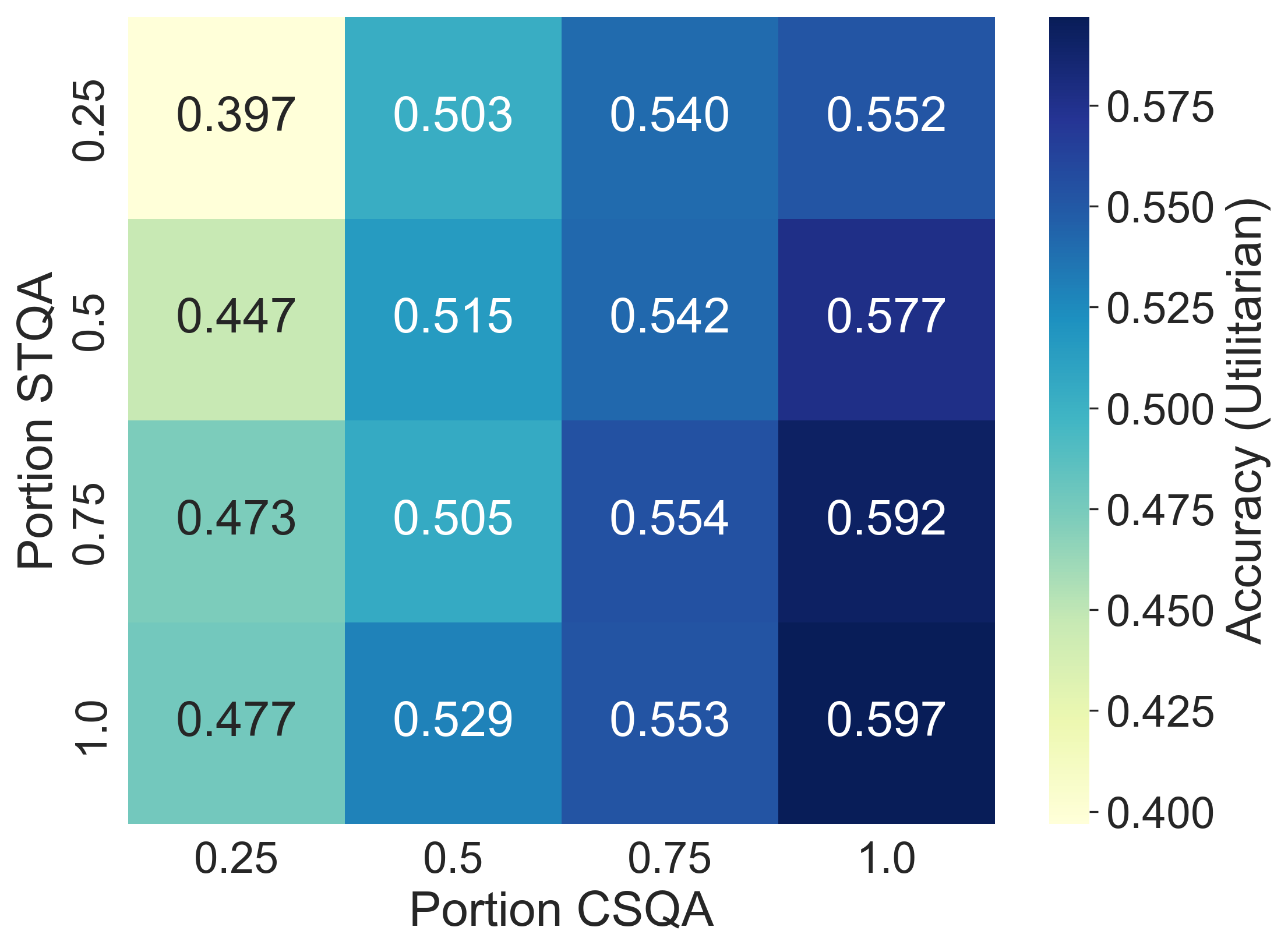}
    \end{subfigure}
    \hfill
    \begin{subfigure}[t]{0.46\linewidth}
        \centering
        \includegraphics[width=\linewidth]{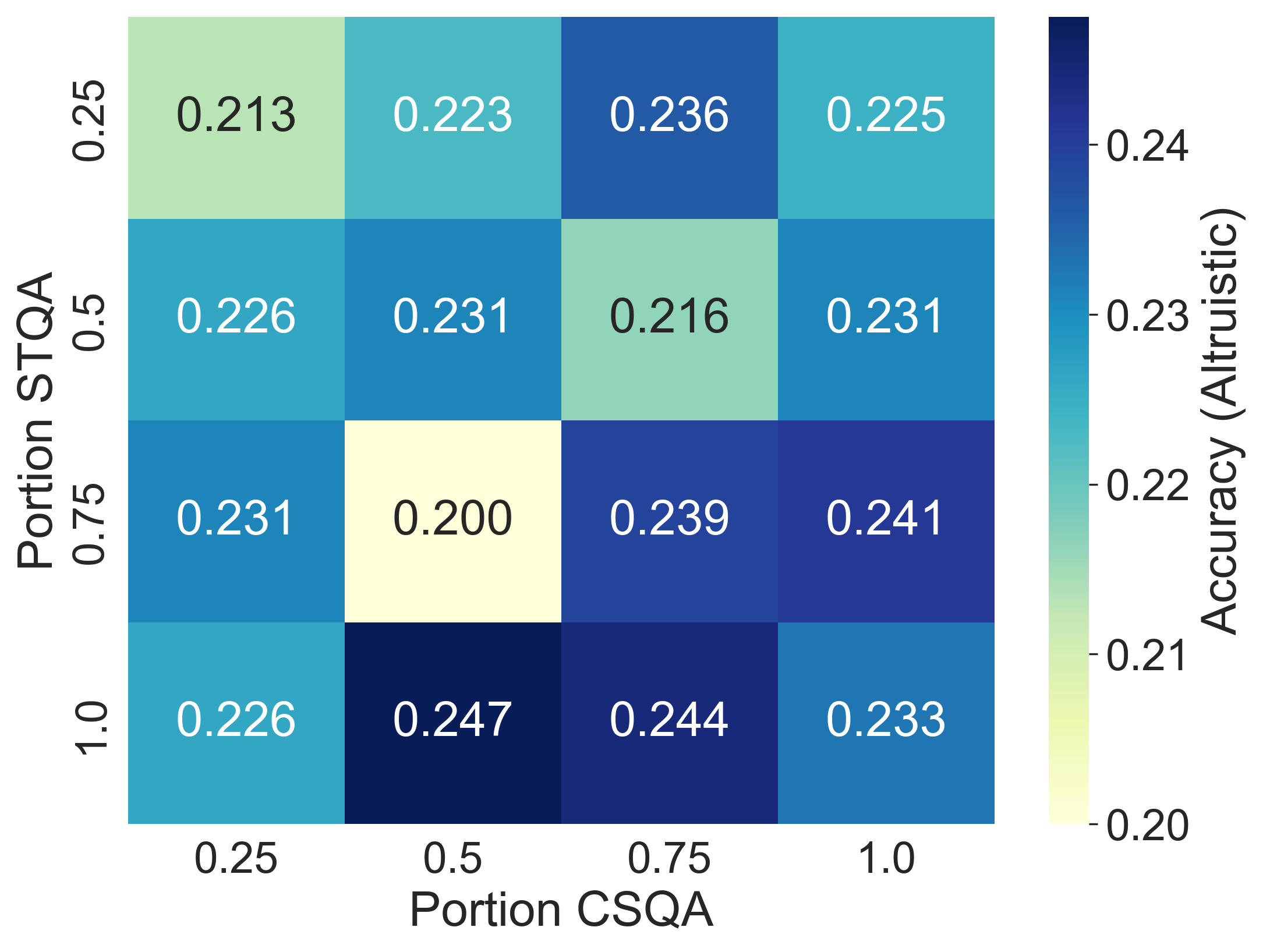}
    \end{subfigure}
    \caption{Impact of varying CSQA and STQA proportions on utilitarian accuracy (left) and altruistic accuracy (right) under full CoT setting (level 6).}
    \label{fig:rq4}
\end{figure}

To explore this dynamic, we vary the contribution proportions of CSQA and STQA across all possible combinations of $\{0.25, 0.5, 0.75, 1.0\}$, resulting in $16$ experiments under the full CoT setting (level 6). \autoref{fig:rq4} visualizes the results as a heatmap. It reveals a clear pattern: utilitarian accuracy generally increases as the contribution from CSQA and STQA grows. However, it is not symmetric between datasets, but it improves more rapidly when the portion of CSQA increases compared to when STQA dominates. This suggests that CSQA contributes disproportionately to performance gains in the utilitarian setting, likely due to its richer multiple-choice structure that provides stronger supervision. In contrast, the right heatmap shows altruistic accuracy, where no such monotonic trend emerges. Instead, performance fluctuates across different ratios, indicating that altruistic outcomes are less sensitive to dataset size increases and more influenced by the interaction balance between CSQA and STQA.

\section{Summarization of Chains} \label{appendix:summarize_chains}

We generate summarized reasoning chains from detailed ones to better reflect real-world constraints, such as contributors limiting detail for reasons of privacy or efficiency.
Specifically, we use Level 6 reasoning traces, which provide the most comprehensive and detailed breakdown of the thought process.
The summarization is performed with two LLMs of different sizes: \texttt{Llama-3.1-8B-Instruct} and \texttt{Llama-3.1-70B-Instruct}.
The models were tasked with condensing the original verbose reasoning into a concise, high-level rationale.
The specific prompt used to guide the summarization process is detailed in \autoref{fig:prompt_rewrite}.

\begin{figure}[h]
    \lstset{style=promptstyle}
    \begin{lstlisting}
System message:
You are an expert at analyzing and summarizing step-by-step reasoning.

User instruction:
Instruction: {instruction}
Original Reasoning: {output}
Rewrite the 'Original Reasoning' into a concise, high-level rationale in |$\leq$|280 characters, including minimal internal steps if necessary. NOTE: When concluding the answer, keep the sentence structure identical to the original reasoning; for example, if the original answer is written as 'The answer is [option]', then the final answer should follow the same format. Do not add any section headers, commentary, or meta-text - just provide the streamlined reasoning followed by the conclusion.
    \end{lstlisting}
    \caption{Prompt used for reasoning summarization with system and user instructions.}
    \label{fig:prompt_rewrite}
\end{figure}

\section{Reasoning Trace Granularity} \label{appendix:level_granularity}
\citet{chen2025unveilingkeyfactorsdistilling} define reasoning trace granularity as the level of detail included in the reasoning process provided by a model. Higher levels of granularity contain detailed step-by-step deductions, intermediate computations, and multiple verification steps, while lower levels include only the essential reasoning and conclusion. 

To illustrate, consider the following question:

\begin{quote}
\textit{If $x$ and $y$ are positive odd integers, then which of the following must also be an odd integer? \\
I. $x^{(y+2)}$ \quad II. $x(y+3)$ \quad III. $(y+1)^{(x-1)} + 1$ \\
Answer Choices: (A) I only \quad (B) I,III only \quad (C) II only \quad (D) III only \quad (E) II,III only}
\end{quote}

In the following, we show two reasoning traces for the same question at different levels of granularity.
\paragraph{Level 1 (Highly Concise Reasoning).}
\begin{quote}\small
\texttt{
Since $x$ is odd, $x^{(y+2)}$ is also odd. Also, $(y+1)^{(x-1)} + 1$ is odd because $y+1$ is even. The answer is B.
}
\end{quote}

This trace compresses the reasoning into a minimal justification of the conclusion without intermediate steps or verification.

\paragraph{Level 6 (Highly Detailed Reasoning).}
\begin{quote}\small
\texttt{
Let's delve into each option with utmost detail: \\
I. $x^{(y+2)} = x^{(2m+1+2)} = x^{(2(m+1)+1)} = (2k+1)^{(2m+1)} = (2k+1)^{(2m)} * (2k+1) = (even+1)^{(even)} * odd$. To understand this better, let's consider the parity of each component: even+1 is odd, even is even, and odd is odd. So, even * odd = odd. We can then verify this by plugging in values, e.g., $x=3, y=5$, then $x^{(y+2)} = 3^{(5+2)} = 3^7 = 2187$, which is odd. We can also consider other cases, such as $x=1, y=7$, or $x=5, y=3$. In all cases, $x^{(y+2)}$ is odd. \\
II. $x(y+3) = x(2m+3) = (2k+1)(2m+3) = 4km + 6k + 2m + 3 = even + even + odd = even$. Again, let's verify with $x=3, y=5$, then $x(y+3) = 3(5+3) = 3*8 = 24$, which is even. We can also try other cases, such as $x=1, y=7$, or $x=5, y=3$. In all cases, $x(y+3)$ is even. \\
III. $(y+1)^{(x-1)} + 1 = (2m+1)^{(2k-1)} + 1 = (even+1)^{(even-1)} + 1 = even^{(even-1)} + 1 = even + 1 = odd$. We can verify this with $x=3, y=5$, then $(y+1)^{(x-1)} + 1 = (5+1)^{(3-1)} + 1 = 6^2 + 1 = 36 + 1 = 37$, which is odd. We can also consider other cases, such as $x=1, y=7$, or $x=5, y=3$. In all cases, $(y+1)^{(x-1)} + 1$ is odd. The answer is B.
}
\end{quote}

This trace includes algebraic derivations, parity reasoning, and multiple numerical checks, providing a fully transparent step-by-step explanation.

\section{Measuring Community Diversity} \label{sec:diversity}
In the context of CDC, where data originate from members of multiple communities, understanding how textual diversity within and between communities affects collective learning is important. The Vendi Score (von Neumann diversity), introduced by \citet{friedman_vendi_2023}, quantifies dataset diversity as the exponential of the Shannon entropy of the eigenvalues of a similarity matrix. It provides a reference-free measure and supports user-defined similarity functions. A high Vendi Score indicates that the data span a wide range of semantic domains, while a low score suggests homogeneity. 

To assess how combining the datasets affects diversity, we compute a diversity gain metric. Let $\{D_1,\dots,D_n\}$ denote datasets (e.g., \texttt{aqua}, \texttt{csqa}, \texttt{obqa}, \texttt{stqa}).  
For each data set $D_i$, we calculate the Vendi Score $\mathrm{VS}(D_i)$ using the SimCSE embeddings \citep{gao_simcse_2022} (\texttt{princeton-nlp/unsup-simcse-bert-base-uncased}). Given a reference dataset $D_i$ and a candidate dataset $D_j$, we define the merge-gain matrix $G$ as
\begin{equation*}
    G_{i,j} =
  \begin{cases}
    \mathrm{VS}(D_i), & i=j, \\[6pt]
    \dfrac{\mathrm{VS}(D_i \cup D_j) - \mathrm{VS}(D_i)}{\mathrm{VS}(D_j)}, & i \neq j.
  \end{cases}
\end{equation*}
Here, $\mathrm{VS}(D_i \cup D_j)$ denotes the Vendi Score computed on the union of the questions from $D_i$ and $D_j$. The off-diagonal term $G_{i,j}$ captures the normalized diversity gain of augmenting $D_i$ with $D_j$. Specifically, the numerator measures the absolute increase in diversity when $D_j$ is added to $D_i$, while the denominator normalizes this increase by the diversity of $D_j$. A positive $G_{i,j}$ indicates that merging $D_j$ introduces additional, non-redundant diversity beyond what is already present in $D_i$, while a negative value suggests that the two datasets contain overlapping or redundant information.
This asymmetric formulation reflects the perspective of $D_i$ as the reference dataset being enriched by contributions from $D_j$.

Based on \autoref{tab:merge_gain_matrix}, it can be observed that datasets containing open-ended questions, such as \texttt{stqa}, which exhibit high internal diversity, do not benefit from additional diversity when merged with other datasets. In contrast, low-diversity datasets gain substantially more diversity when combined with high-diversity datasets. 

\begin{table*}[ht]
\centering
\begin{tabular}{lcccc}
\toprule
 & aqua & csqa & obqa & stqa \\
\midrule
aqua & $\mathrm{VS}(\texttt{aqua})$ = 35.73 & 0.3091 & 0.3349 & 0.4310 \\
csqa & 0.3698 & $\mathrm{VS}(\texttt{csqa})$ = 32.59 & 0.1995 & 0.4151 \\
obqa & 0.3117 & 0.0917 & $\mathrm{VS}(\texttt{obqa})$ = 36.97 & 0.4038 \\
stqa & -0.5546 & -0.7519 & -0.5738 & $\mathrm{VS}(\texttt{stqa})$ = 97.61 \\
\midrule
\end{tabular}
\caption{Merge–gain matrix using SimCSE embeddings. Rows correspond to the reference dataset, and columns correspond to the candidate dataset. Diagonal entries represent single-dataset Vendi scores.}

\label{tab:merge_gain_matrix}
\end{table*}


\end{document}